\documentclass[10pt,journal,compsoc]{IEEEtran}

\ifCLASSOPTIONcompsoc
  \usepackage[nocompress]{cite}
\else
  \usepackage{cite}
\fi

\usepackage[numbers]{natbib}
\usepackage{soul}
\usepackage{multirow}
\usepackage[table]{xcolor}

\ifCLASSINFOpdf
  \usepackage[pdftex]{graphicx}
\else
  \usepackage[dvips]{graphicx}
\fi

\usepackage{xcolor}

\usepackage{algorithm}
\usepackage{algpseudocode}

\usepackage{threeparttable}
\usepackage{subcaption}
\usepackage{amsthm}
\usepackage{amssymb}
\usepackage{amsmath}

\usepackage{float}

\theoremstyle{definition}
\newtheorem{definition}{Definition}

\newtheorem{theorem}{Theorem}

\newtheorem{lemma}{Lemma}

\usepackage{hyperref}
\usepackage{enumitem}

\begin{document}

\title{Personalized Privacy-Preserving Framework\\ for Cross-Silo Federated Learning}

\author{Van-Tuan Tran,
        Huy-Hieu Pham, \textit{Member, IEEE}
        and~Kok-Seng Wong*, \textit{Member, IEEE}
\IEEEcompsocitemizethanks{\IEEEcompsocthanksitem 
*Corresponding author: Kok-Seng Wong (\href{mailto:wong.ks@vinuni.edu.vn}{wong.ks@vinuni.edu.vn})
\IEEEcompsocthanksitem Hieu H. Pham and Kok-Seng Wong are with the College of Engineering and Computer Science (CECS), VinUniversity, Hanoi, Vietnam.
\IEEEcompsocthanksitem Van-Tuan Tran, Hieu H. Pham, and Kok-Seng Wong are with the VinUni-Illinois Smart Health Center (VISHC), VinUniversity, Hanoi, Vietnam.}%
}

\markboth{Journal of \LaTeX\ Class Files,~Vol.~14, No.~8, August~2015}%
{Shell \MakeLowercase{\textit{et al.}}: Bare Demo of IEEEtran.cls for Computer Society Journals}

\IEEEtitleabstractindextext{%
\begin{abstract}
Federated learning (FL) is recently surging as a promising decentralized deep learning (DL) framework that enables DL-based approaches trained collaboratively across clients without sharing private data. However, in the context of the central party being active and dishonest, the data of individual clients might be perfectly reconstructed, leading to the high possibility of sensitive information being leaked. Moreover, FL also suffers from the nonindependent and identically distributed (non-IID) data among clients, resulting in the degradation in the inference performance on local clients’ data. In this paper, we propose a novel framework, namely Personalized Privacy-Preserving Federated Learning (PPPFL), with a concentration on cross-silo FL to overcome these challenges. Specifically, we introduce a stabilized variant of the Model-Agnostic Meta-Learning (MAML) algorithm to collaboratively train a global initialization from clients' synthetic data generated by Differential Private Generative Adversarial Networks (DP-GANs). After reaching convergence, the global initialization will be locally adapted by the clients to their private data. Through extensive experiments, we empirically show that our proposed framework outperforms multiple FL baselines on different datasets, including MNIST, Fashion-MNIST, CIFAR-10, and CIFAR-100. The source code for this project is available at \href{http://github.com/github.com/vinuni-vishc/PPPF-Cross-Silo-FL}{https://github.com/vinuni-vishc/PPPF-Cross-Silo-FL}.

\end{abstract}
}

\maketitle

\IEEEdisplaynontitleabstractindextext

%
\IEEEpeerreviewmaketitle

\IEEEraisesectionheading{\section{Introduction}\label{sec:introduction}}

\IEEEPARstart{A}{lthough} deep learning (DL) approaches have achieved significant accomplishments in a wide range of applications \cite{suvorov2022resolution, saharia2022photorealistic, schulman2022chatgpt}, there exists several fundamental challenges \cite{dong2021survey} that yet need to be addressed in order to enable its complete power. Due to data sovereignty and data privacy concerns, one major problem is the difficulty in the centralization of the local data from multiple clients to one central server. 
In order to guarantee the privacy-preserving in training schemes, federated learning (FL) framework \cite{mcmahan2017communication} is proposed to allow multiple entities (e.g., individuals, or organizations) collaboratively train a DL model under the coordination of a central server. Specifically, each client's data is trained and stored locally, and only updated gradients are transferred to the server for aggregation purposes. By keeping the training data locally and in a decentralized fashion, FL approaches can alleviate many of the systemic privacy risks of traditional centralized DL methods \cite{zhang2021survey}. 

However, under FL's naive implementation, there are several key challenges \cite{yang2019federated}, including private information leakage, verifiable computation, expensive communication costs, and statistical heterogeneity. 
In the scope of this research, we focus on the problem of private information being divulged in FL settings. Generally, despite the local training without data exchange between the server and clients, private information can still be leaked to some extent from the local gradients through inversion attacks \cite{geiping2020inverting,wang2019beyond,yin2021see,zhao2020idlg,zhu2019deep}. More seriously, as the central server holds full control of observing and modifying the shared global model, sensitive information can also be eavesdropped on by the server. For instance, \citeauthor{boenisch2021curious} \cite{boenisch2021curious} identified that an active-and-dishonest server can set up trap weights for recovering the clients' data with zero reconstruction loss. 

With the aim of limiting information disclosure about private data, a state-of-the-art privacy-preserving technique was proposed by \citeauthor{dwork2014algorithmic} \cite{dwork2014algorithmic}, known as differential privacy (DP). The central principle of this technique is to add curated noises to the computational steps to perturb the sensitive information. 
In general, there are two approaches to incorporating DP into FL settings: global DP \cite{dwork2006calibrating} and local DP \cite{duchi2013local}. Under global DP, clients are assumed to trust the centralized server, and noises are added by the server; while under local DP, each client perturbs its information locally prior to sharing it with the server without requiring trust in a server \cite{kasiviswanathan2011can}. Despite the addition of noises, global DP approaches cannot ensure privacy preservation in the case of the server being dishonest. Therefore, in this research, we concentrate on the local DP scenario to enhance the privacy protection of the clients against gradient leakage and parameter modification.

Another issue in FL is the non-independent and identically distributed (non-IID) \cite{zhu2021federated, kulkarni2020survey}, caused by the heterogeneity in local data distributions among clients. In particular, in non-IID scenarios, different clients train models from different data distributions, which leads to the performance degradation of the aggregated model. This problem was especially taken into consideration in the cases of cross-silo FL  where clients are large entities (e.g., companies, institutions, and hospitals), and their private data may be greatly different in size and partitioning \cite{kairouz2021advances}. There are plenty of approaches introduced for handling non-IID problems, such as sharing data with the server \cite{tuor2021overcoming}, fine-tuning the local models after obtaining the aggregated model \cite{wang2019federated, t2020personalized}, adding personalized layers in local models \cite{arivazhagan2019federated}, and transferring knowledge from the server to a certain client \cite{liu2020secure, lin2020ensemble}. While those methods mitigate the non-IID data distribution in FL in some cases, they inevitably increase the risk of privacy leakage and a large number of hyperparameters \cite{zhu2021federated}.


\subsection{Our Contributions}

In this work, to mitigate the above challenging FL problems, we propose a Personalized Privacy-Preserving Federated Learning (PPPFL) framework for cross-silo FL, a novel framework that allows local clients simultaneously protect their own initial data at the instance-level and achieve a personalized model. To the best of our knowledge, this study is the first to marry two lines of research in FL on privacy-preserving and non-IID data, to achieve both differential privacy guarantees and better convergence performance. This framework consists of two computational stages: server-side computations and client-side computations. In terms of the former stage, building on meta-learning methods, the primary goal is to find an initial global model shared between all clients which performs well after each client updates it with respect to its own generated artificial data (Section \ref{sec:server-compute}). In this way, the final model implemented by each local client can achieve stronger generalization guarantees and more personalized adaptation. Regarding the latter stage, each local client generates synthetic data by training a differentially private data generative model (Section \ref{sec:dp-training}). 
Those data are then utilized for collaboration purposes in the Federated training phase, therefore preventing sharing of the raw data (Section \ref{sec:client-compute}). Moreover, we also focus on cross-silo FL as attention put on applying FL to multiple large organizations has greatly grown \cite{cordis2019machine, treleaven2022federated}. Therefore, it is sufficient for our proposed GANs-based framework to guarantee the DP in cross-silo FL.
\\

We summarize our contributions as follows:

\begin{itemize}
    \item We propose a PPPFL, a novel framework that not only prevents information leakage but also tackles the non-IID issue. To the best of our knowledge, we are the first to introduce a new approach to provide the instance-level privacy-preserving and address data heterogeneity in FL. 
    
    \item We first consolidate several advantages of \textsc{DataLens} into a single FL framework to provide instance-level DP, helping to minimize the risks of sensitive information leakage.

    \item We introduce a new variant of FedMeta in terms of personalization and convergence by incorporating the clients' privacy budgets and gradient stabilization techniques into the optimization process.
    
    \item We conduct thorough experiments and ablation studies to demonstrate the effectiveness of the proposed approach on different datasets.\\
    
\end{itemize}
\leavevmode
\noindent  The rest of this paper is organized as follows. Related works on privacy-preserving FL and non-IID challenges in FL are reviewed in Section \ref{sec:related}. In section \ref{sec:preliminary}, we introduce backgrounds on threat models, differential privacy, data generative models, and optimization-based meta-learning methods. Section \ref{sec:method} first presents our problem formulation and details of the proposed method, then performs a rigorous analysis of its privacy guarantee and convergence. Section \ref{sec:evaluation} provides extensive experiments on multiple benchmark datasets. Finally, section \ref{sec:discussion} discusses the experimental results, some key findings, and limitations of this work and concludes the paper.

\section{Related Works}
\label{sec:related}
\noindent
\textbf{Privacy-preserving FL.} 
Because of the Active-and-dishonest threat server, in this work, we mainly focus on local DP in which noise addition is conducted by each local client to prevent perfect extractability of individually private information. Many DP approaches based on the SGD algorithm have been proposed to protect the privacy of clients. \citeauthor{sun2020ldp} \cite{sun2020ldp} used the adaptive range of clients' weights to achieve better-aggregated model performance, especially in deeper models, and leveraged parameter shuffling as an extra layer of privacy protection. Meanwhile, \cite{truex2020ldp} allowed the individual clients to define their own local privacy setting by developing a local training approach and a parameter update sharing method, based on Condensed Local Differential Privacy with Exponential Mechanism \cite{gursoy2019secure}. In order to minimize the tradeoff between the privacy budget and the convergence performance, \citeauthor{wei2021user} \cite{wei2021user} proposed a DP-SGD-based communication rounds discounting method.

Another approach to mitigating the privacy concern for data sharing is to utilize generative models. The key idea of this approach is consolidating generative models to locally synthesize data, and then these generated data would be sent to the server by clients for collaboration. \citeauthor{chen2018differentially} \cite{chen2018differentially} first leveraged the power of autoencoder in FL that allows clients to generate their own artificial data by independently feeding their private data into the autoencoder. Under the FL framework and DP thinking, \cite{zhang2021feddpgan} introduced FedDPGAN for COVID-19 pneumonia detection without compromising patients' sensitive information. In this method, each medical institution adds Gaussian noise in training gradients, and then parameters are sent to the server for training COVID-19 diagnostic models collaboratively.

However, those approaches have not taken non-IID data into consideration which inevitably causes a deterioration in the accuracy of FL. In this research, we not only handle privacy concerns but also focus on non-IID problems. Moreover, in real-world industrial environments which is referred to as the cross-silo setting, \cite{mahawaga2022local} utilized RAPPOR \cite{erlingsson2014rappor} and optimized unary encoding that can efficiently control the privacy budgets and address the massive data distribution in many industrial applications (e.g., healthcare, smart cities). Inspired by this work, we also provide a framework for privacy-preserving FL to be utilized under cross-silo settings, potentially with untrusted parties.\\
\leavevmode
\\
\noindent \textbf{Solving the non-IID problem in FL.} Due to the heterogeneity in participants' data distributions, possessing a global model that can fit all the models among the parties is difficult \cite{kairouz2021advances}. To handle this issue, personalization approaches have received much attention recently. The early work \cite{smith2017federated} cast the personalization problem as a multi-task learning problem, which generates separated but associated models for each client. Meanwhile, \cite{arivazhagan2019federated, liang2020think} allowed each client to have personalized layers followed by base layers, and only the base layers are shared with the parameter server. This approach is able not only to boost the learning performance of FL on unknown heterogeneous data but also to reduce the communication costs since clients are not necessarily to share the whole model with the server. Knowledge distillation is also a promising approach for personalized FL with the main idea of transferring knowledge from the server or other clients to a particular client to enhance its performance on non-IID data. Transfer learning is utilized by \cite{liu2020secure, lin2020ensemble}, in which \citeauthor{liu2020secure} incorporated additively homomorphic encryption (HE) and secret sharing while \citeauthor{lin2020ensemble} leveraged an ensemble distillation strategy respectively, to allow knowledge to be shared without compromising user privacy. Another important chart of knowledge distillation is federated domain adaptation which aims at eliminating the differences between data shards between clients \cite{li2019fedmd, peng2019federated}. The most effective and powerful personalization method is to incorporate meta-learning within FL. MAML \cite{finn2017model} has been leveraged in  \cite{fallah2020personalized}, where authors proposed Personalized FedAvg to seek a proper initialization model for the clients that can be easily updated with the local data of each client after the training phase.

Motivated by meta-learning approaches, we also take advantage of MAML to provide personalized models for clients with non-IID data but introduce a stabilized version of MAML that is more easily convergent compared to that in \cite{fallah2020personalized}. Similar to studies \cite{huang2021personalized, shamsian2021personalized}, our work also emphasizes on cross-silo FL to address challenges in real-world settings, such as high coordination and operational cost \cite{kairouz2021advances}.


\section{Preliminaries}
\label{sec:preliminary}
In this section, we provide the details of threat models in privacy when it comes to the existence of an active-and-dishonest server. Next, we provide some background knowledge on DP and the local DP guarantees in FL. Finally, we briefly discuss Optimization-based Meta-learning through the MAML \cite{finn2017model} framework. We also summarize the crucial notations used in this paper in Table \ref{tab:notations}.

\begin{table}[H]
\caption{Main notations used in the paper.}
\resizebox{\columnwidth}{!}{%
\begin{tabular}{l|l}
\hline
$\theta$                        &   Global model's parameters               \\ \hline
$\theta_i$                      &   Local model's parameters in client $i$  \\ \hline
$D_i$                           &   Secret (private) dataset in client $i$            \\ \hline
$\mathcal{G}_i$                 &   Local DP data generator of client $i$ trained from $D$ \\ \hline
${D^*}_i$                &   Synthetic dataset of client $i$ generated by $\mathcal{G}_i$\\ \hline
$\epsilon_i$                    &   Privacy budget chosen by client $i$ when training $\mathcal{G}_i$       \\ \hline
$\alpha^{*}$                    &   Inner learning rate when training local model $\theta$ using synthetic data ${D^*}$   \\ \hline
$\beta^{*}$                     &   Outer learning rate when server aggregate local models $\theta$ \\ \hline
\end{tabular}
}
\label{tab:notations}
\end{table}

\subsection{Threat models}
\textbf{Semi-honest server.} In this type of server, all participated parties follow the protocol specification to perform the computation task but they may try to infer the data of other parties from the received intermediate computation results \cite{li2021trustworthy}.

\citeauthor{geiping2020inverting} indicated that through data recovery attacks \cite{7958568}, the gradients of any fully connected (FC) layers followed by a ReLU activation function $\mathbf{y} = \textsc{ReLU}(W\mathbf{x} + \mathbf{b})$ contain the training data point $\mathbf{x}$ that may result in the possibility of the original data being reconstructed from the knowledge of the parameter gradients.
Formally, when the $i^{th}$ neuron of a FC layer is fired (i.e., $\mathbf{w}_i^T \mathbf{x} + b_i > 0$), the data point $\mathbf{x}$ can be reconstructed by computing $\mathbf{x}^{T}=\left(\frac{\partial \mathcal{L}}{\partial b_{i}}\right)^{-1} .  \left(\frac{\partial \mathcal{L}}{\partial \mathbf{w}_{i}^{T}}\right)$, where $\mathbf{w}_i^T$ be the  $i^{th}$ row in the weight matrix and $b_i$ be the  $i^{th}$ component in the bias vector.
 Consequently, a passive attacker (e.g., semi-honest server) can extract roughly 20\% of arbitrary data points from a batch size $B =  100$ for 1000 neurons on ImageNet \cite{7958568}.\\
\\
\noindent \textbf{Active-and-dishonest server.}  
Due to the existence of the data leakage through gradients of the FC layer, \citeauthor{boenisch2021curious} has raised high privacy-related concerns in FL settings as the central party holds full control of observing and modifying the shared gradients. Although the addition of the calibrated noise into the users' data or shared parameters can prevent the data reconstruction attack \cite{wei2020federated, adnan2022federated, ramaswamy2020training}, these implementations provide virtually no protection against a non-honest server since the additional step is performed by the server but not the clients themselves. Therefore, it is evident that for the active-but-dishonest server, no privacy guarantees can be provided to the clients.
In this paper, our goal is to design a differentially private framework, which ensures that the synthetic data instead of the sent parameters are differentially private. By this setting, clients only send the model's parameters trained with DPGAN-by generated data to the server for collaboration; thus differential privacy can be guaranteed even in cases of a non-honest central party.

\subsection{Differential Privacy}
$(\epsilon-\delta)$ Differential Privacy proposed by \citeauthor{dwork2014algorithmic}\cite{dwork2014algorithmic} is currently a strong criterion for privacy preservation of distributed data processing systems. It provides a guarantee that adversaries cannot infer sensitive information about individuals based on the algorithm outputs. Some formal definitions and concepts of types of DP in general and in FL are as follows.

\begin{definition}
    \label{def:dp}
    $(\epsilon-\delta)$ Differential Privacy 

    Given a randomized mechanism $M$ and adjacent datasets $D$ and $D'$, $M$ satisfies $(\epsilon, \delta)$-differential privacy, if for all outputs $O$ of $M$ and for all possible sets $S$ of outputs

    \begin{equation*}
    \begin{array}{rl}
        \label{eq:epsilon_delta_dp}
        Pr[M(D) \in S] &\leq \exp(\epsilon) * Pr[M(D') \in S] + \delta.   
    \end{array}
    \end{equation*}
    
    This definition captures the idea that the probability of any particular output should not be significantly affected by the presence or absence of any individual's data in the general dataset. The parameters $\epsilon$ and $\delta$ control the strength of the privacy guarantee, with smaller values of $\epsilon$ and $\delta$ indicating a higher degree of privacy preservation.    
        
\end{definition}

\begin{definition}
    \label{def:rdp}
    $(\lambda, \alpha)$-Renyi Differential Privacy
    
    A randomized mechanism ${M}$ guarantees $(\lambda, \alpha)$-RDP with $\lambda>1$, if for any neighboring datasets $D$ and $D'$
    
    \resizebox{0.95\columnwidth}{!}{
    $D_\lambda\left({M}(D) | {M}\left(D^{\prime}\right)\right)=\frac{1}{\lambda-1} \log \mathbb{E} _ {x \sim {M}(D)} \left[\left(\frac{\operatorname{Pr}[{M}(D)=x]}{\operatorname{Pr}\left[{M}\left(D^{\prime}\right)=x\right]}\right)^{\lambda-1}\right] \leq \alpha.$
    }    
    \end{definition}

\begin{theorem} (From RDP to DP \cite{mironov2017renyi})
    
    If a mechanism $\mathcal{M}$ guarantees $(\lambda, \alpha)$-RDP, then it also guarantees $\left(\alpha+\frac{\log 1 / \delta}{\lambda-1}, \delta\right)$-differential privacy for any $\delta \in(0,1)$.
    
\end{theorem}


\noindent
\textbf{Local DP and Global DP.}
In FL, global differential privacy (global DP) refers to a privacy-preserving technique in which a trusted aggregator adds noise to multiple users' data or parameter updates to protect their privacy. The noise is added using a DP mechanism, which is designed to ensure that the noise is sufficient to prevent the identification of any individual user's data, even when the aggregator has access to the data of all users.
However, when the central server, or aggregator, is active and dishonest, it could identify the data of individual users. Hence, local DP adds noise to the data locally before it is sent to the central server or transmitted over an insecure network.



\textbf{User-level DP and instance-level DP in FL.}
User-level DP is a variant of local DP used when the data consists of a smaller number of samples associated with specific users, and the focus is on protecting the privacy of the users rather than the privacy of the individual data points.

On the other hand, instance-level differential privacy is generally more robust than user-level DP because it provides stronger privacy guarantees for individual data samples. Particularly, instance-level DP ensures that the probability of an adversary learning sensitive information about a specific data point is bounded by a small privacy budget. Moreover, instance-level DP is  easier to implement and apply in practice than user-level DP, as it does not require tracking or identifying specific users or user groups in the dataset, thus making it more flexible and easier to use, especially in the context of cross-silo FL \cite{liu2020learning, wei2021user, levy2021learning}.

\subsection{Differentially Private Data Generation} 

Generative Adversarial Networks (GANs) are generative models that consist of a generator and a discriminator, which are trained to play a minimax game to generate synthetic data that is similar to a given dataset. 
More formally,  the generator $\mathcal{G}$ maps from a noise variable $z \sim p_z(z)$ to a fake sample $\mathcal{G}(z)$ and to trick the discriminator $\mathcal{D}$ from classifying fake sample against real sample $x \sim p_r(x)$. 
The loss function can be formulated as

\begin{equation*}
\label{eq:gan}
\begin{split}
    \min_\mathcal{G} \max_\mathcal{D} L(\mathcal{D}, \mathcal{G})  &= \mathbb{E}_{x \sim p_{r}(x)} [\log \mathcal{D}(x)] \\
    &\quad + \mathbb{E}_{z \sim p_z(z)} [\log(1 - \mathcal{D}(\mathcal{G}(z)))].
\end{split}
\end{equation*}
    

GANs are successfully utilized in plenty of applications \cite{wang2021generative}, such as image in-painting, image super-resolution, image-to-image translation, language, and speech synthesis.
Due to the high expressiveness of deep neural networks in modeling the data distribution, DL-based generative models can output high-fidelity samples. However, similar to other DL-based architectures, GAN models are also vulnerable to several privacy violation attacks \cite {chen2018differentially} that might extract massive sensitive information from generated data. In order to provide rigorous privacy guarantees to synthetic data, many approaches adopted DP in GANs in hopes of enabling high learning efficacy without disclosing much sensitive information about individuals within the training set. DP-GAN \cite{xie2018differentially} and DP-CGAN \cite{torkzadehmahani2019dp} are two earlier approaches introduced to adapt DP-SGD \cite{abadi2016deep} to GAN. The principle of this method is to add calibrated Gaussian noise to the discriminator gradients during the training step. Later, in order to address a major limitation of prior studies related to privacy budget explosion, \citeauthor{papernot2018scalable} \cite{jordon2018pate} integrated the off-the-shelf Private Aggregation of Teacher Ensembles (PATE) framework \cite{papernot2018scalable} into GAN to provide better privacy budget usage while mitigating the sensitive data disclosure. Specifically, PATE-GAN first learns a set of teacher discriminators on a disjoint subset of training data and then aggregates the output of these teacher models to train a student discriminator. Another variant, G-PATE \cite{long2019scalable} improved PATE-GAN by directly training scalable differentially private student generators to preserve high generated data utility. Nevertheless, those still cannot generate differentially private high dimensional data with strong privacy guarantees. By combining the PATE framework with a $top-k$-based proposed noisy gradient compression and aggregation strategy \textsc{TopAgg}, \textsc{DataLens} \cite{wang2021datalens} demonstrated the capability of generating high-dimensional data with limited privacy budgets and communication-efficient distributed learning, outperforming PATE-GAN and G-PATE.
Taking inspiration from DP-GAN, especially \textsc{DataLens}, we can enjoy the benefits of them to synthesize high-utility data while protecting individual training information. We provide a detailed description in Section \ref{sec:client-compute} and \ref{sec:server-compute}.


\subsection{Optimization-based Meta-learning}

The goal of meta-learning is to train a model that can quickly adapt to a new task, based on only a few examples, through two phases: meta-learning and adaptation. To achieve effectiveness, the meta-learning problem treats entire tasks as training samples. Optimization-based approaches cast the adaption part of the process as an optimization problem \cite{finn2017model}. This  aims at establishing initial model parameters $\theta$ through learning from $T$ number of similar tasks, which can perform well on a new task $T+1$. 
Formally, given a task $\mathcal{T}$, and its associated training and validation dataset $(\mathcal{D}^{tr}, \mathcal{D}^{val})$, the optimal parameter $\theta^{*}$ for $\mathcal{D}^{tr}$ is calculated by a gradient descent step
\begin{equation*}
    \theta^{*} = \theta - \alpha\nabla_{\theta}\mathcal{L}(\theta, \mathcal{D}^{tr}),
\end{equation*}
where $\mathcal{L}$ is the loss function, each $\mathcal{D}$ contains observations $X$ and corresponding targets $y$, $\mathcal{D} = \{X, y\}$.
Different from transfer learning, we need to optimize the initial parameter $\theta$ such that it would perform well on the validation set $\mathcal{D}^{val}$. In meta-learning, this step is known as the meta-training step
\begin{equation*}
    \theta = \theta - \beta\nabla_{\theta}\mathcal{L}(\theta^{*}, \mathcal{D}^{val}).
\end{equation*}

For multiple tasks $\mathcal{T} = \{\mathcal{T}_1, \mathcal{T}_2, ..., \mathcal{T}_t\}$, the optimal parameter $\theta^{*}$ for a task $\mathcal{T}_i$ is computed
\begin{equation*}
    \theta^{*}_i = \theta - \alpha\nabla_{\theta}\mathcal{L}(\theta, \mathcal{D}^{tr}_i),
\end{equation*}
with the corresponding updating rule
\begin{equation*}
    \theta = \theta - \beta\nabla_{\theta}\sum_i{\mathcal{L}(\theta^{*}_i, \mathcal{D}^{val}_i)}.
\end{equation*}

A task $t$ contains i.i.d. observations $X_t$ and their corresponding targets $y_t$.  
Similarly, to evaluate a meta-learning algorithm, one usually split each task's data $(X_t, y_t)$ into the training (support) set $(X_t^{tr}, y_t^{tr})$ and validation (query) set $(X_t^{val}, y_t^{val})$. If we define $\mathcal{L}_t: \mathbb{R}^d \rightarrow \mathbb{R}$ as the loss corresponding to task $t$.





\textbf{Meta-training} is to learn a model that can adapt to new tasks quickly and efficiently using only a small number of gradient descent steps, without the need for extensive training data. Formally, the training objective function can be formulated as follows

\begin{equation}
    \label{eq:meta_training}
    \resizebox{0.9\columnwidth}{!}{
    $L^{train}(\theta) = \frac{1}{T}\sum_{t=1}^{T}{\mathcal{L} \left[\left(\theta; X_t^{tr}, y_t^{tr} \right); X_t^{val}, y_t^{val} \right].}$
    }
\end{equation}

\textbf{Meta-testing} is the process of evaluating the performance of a model that has been meta-trained on a set of new tasks. During meta-testing, the model is presented with a novel task that is similar but has not been seen during meta-training. The empirical risk on novel task $T+1$ can be formulated as follows

\begin{equation}
    \label{eq:meta_test}
    \begin{split}
     L^{test}(\theta) &= \mathbb{E}_{(X_{T+1}, y_{T+1}, (x', y') \sim p_{T+1})} \\
     &\quad \mathcal{L}\left[(\theta; X_{T+1}, y_{T+1}); x', y') \right].
     \end{split}
\end{equation}


\begin{figure*}[!ht]
    \centering
    \includegraphics[width=0.75\textwidth]{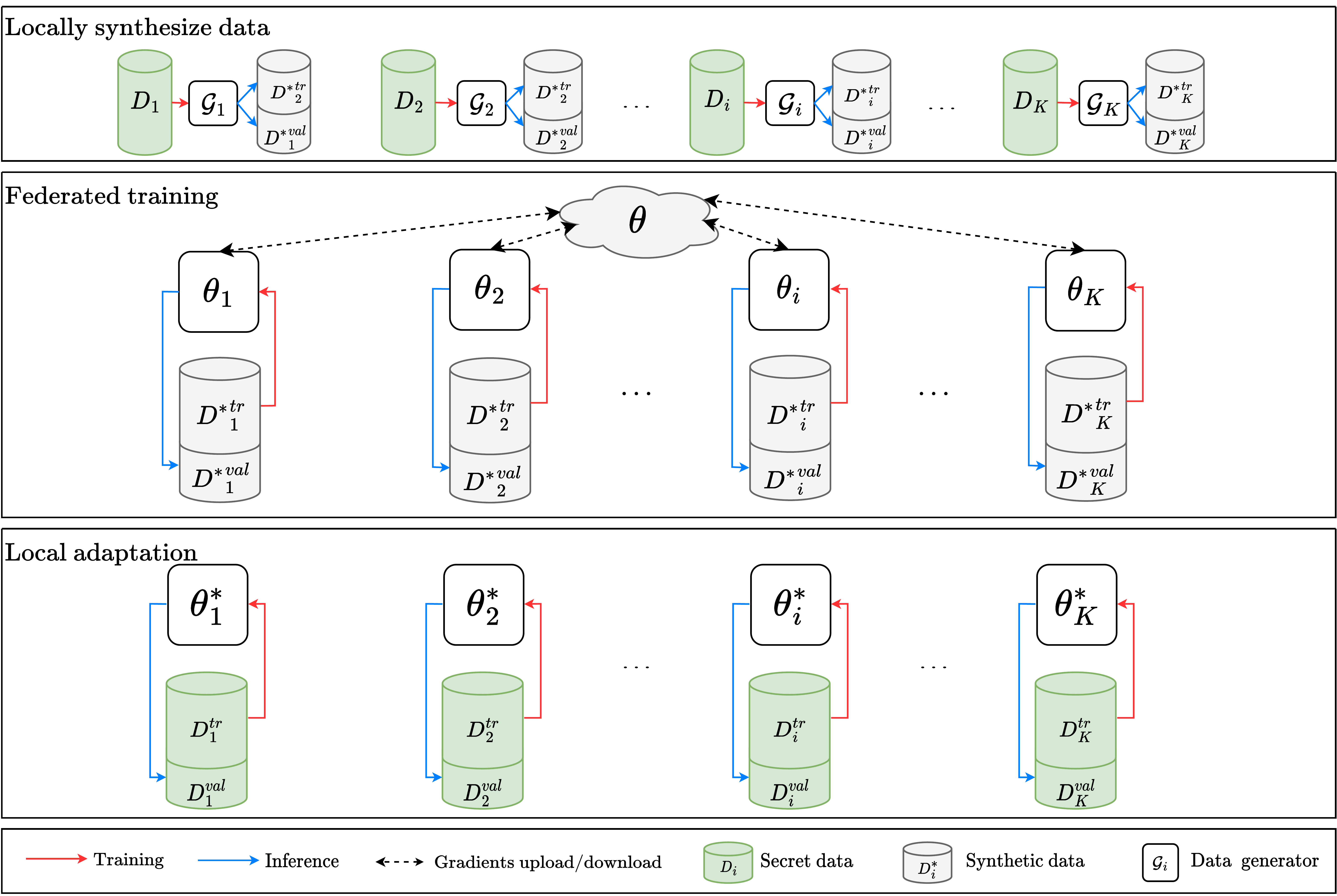}
    \caption{The overall architecture of the PPPFL framework consisting three consecutive stages. First, each participant locally synthesizes DP-guaranteed dataset ${D^*}$ by training their own DP-data-generator $\mathcal{G}$ from secret dataset $D$. Next, in the federated training process, the synthetic data, which have been respectively separated into a training set ${D^*}^{tr}$ and held-out validation set ${D^*}^{val}$, will be used to update the global model parameters $\theta$. Lastly, the clients get emerged from the federated training rounds and perform updates to their local model parameters $\theta^{*}$ based on their secret training data $D^{tr}$ and evaluate their new locally adapted model with respect to the secret validation set $D^{val}$.}
    \label{fig:framework}
\end{figure*}

\section{Methodology}
\label{sec:method}

In this section, we present a formal problem formulation in which the goal is to enhance privacy in personalized Federated Learning. We then proceed to describe our proposed solution to this problem, the PPPFL framework, in detail. This includes a step-by-step description of the process, beginning with the training of local models on the client side and culminating in the federated collaboration between the clients and the server. Finally, we provide theoretical proof demonstrating the privacy and convergence guarantees of the PPPFL framework.

\subsection{Problem Formulation}
\noindent
\textbf{Privacy-preserving FL.} Given a set of $K$ clients $\{ c_1, \dots, c_K \}$ sampled from a client distribution $\mathcal{C}$ and an un-trusted server $S$.
Each $i^{th}$ client has its own private dataset $D_i$ and all clients' datasets are non-IID. In this research, we mainly study the common non-IID scenario where the local data distribution $\Pr(x_i, y_i) = \Pr(x_i \mid y_i) \Pr(y_i)$ with the label distribution $\Pr(y_i)$ is different among clients. 
For the evaluation purpose, local private datasets $D_i$ are divided into training $D_i^{tr}$ and testing $D_i^{val}$ sets, such that $D_i = D_i^{tr} \cup D_i^{val}$. Due to the concern of the active-and-dishonest server $S$, the private datasets will be kept locally and never be used for the federated training phase. Instead, clients train their own personalized  synthetic data generator $\mathcal{G}_i$, and then use the generated datasets ${D^*}_i^{tr}$ and ${D^*}_i^{val}$ with ${D^*}_i = \mathcal{G}_i(D_i)$ for collaboration. The privacy budget $\epsilon_i$ in each $\mathcal{G}_i$ is customized by the clients. 

\noindent
\textbf{Personalized FL.} The goal is to find an initial shared model that can perform well after each client updates it with respect to its own gradients. Specifically, the central party uses synthetic datasets ${D^*}_i$ to meta-learn a generalized model parameterized by $\theta$ from the updates sent by the clients.  Therefore, this model can be later personalized by each client with its own corresponding private data $D_i$.  The support set ${D^*}_i^{tr}$ and a query set ${D^*}_i^{val}$, and the meta-model will be locally adapted to the private data set $D_i^{tr}$ and evaluated on $D_i^{test}$ by the clients.

Considering the meta-train tasks and meta-test tasks $t \sim \mathcal{T}$ are now the clients $i \sim \mathcal{C}$ with the synthetic ${D^*}_i$ and private data $D_i$, respectively. The meta-training objective function in Eq. \ref{eq:meta_training} can be re-written as

\begin{equation}
    L(\theta) = \mathbb{E}_{c \sim \mathcal{C}}\mathcal{L}\left[\left({\theta}; {D^*}_{c}^{tr}\right); {D^*}_i^{val}\right].
\end{equation}

\subsection{Our Proposed Framework}
We first introduce the data flow in each client and the central server. Then, we present our techniques to overcome the challenges of gradients instability \cite{antoniou2018train} when using MAML \cite{finn2017model} as the optimization-based meta-learning algorithm, and the computational steps of the \textsc{DataLens} \cite{wang2021datalens} architecture. 

\begin{algorithm}[H]
    \caption{\textsc{Computational steps on the clients}} 
    \label{alg:clients}
    \begin{algorithmic}[1]
    \renewcommand{\algorithmicrequire}{\textbf{Input:}}
    \renewcommand{\algorithmicensure}{\textbf{Output:}}
    \renewcommand{\algorithmiccomment}[1]{\hfill// #1}
    \Require client data $D$, client model $\theta$, fine-tuning steps $\gamma$
    \Ensure  adapted model $\hat{\theta}$
    \\ \textit{// Before collaborative training}
    \For{each client $c \in \mathrm{C}$}
        \State $\mathcal{G}_{c} \gets \textsc{DP-GAN Training}(D_{c}$) \ref{alg:datalens} \Comment{Training the DP data generator}
        \State (i) $D_{c} \gets D^{tr}_{c} \cup D^{val}_{c}$ \Comment{Locally data partitioning}
        \State (ii) ${D^*}^{tr}_{c} \gets \mathcal{G}_{c}(D^{tr}_{c}),   {D^*}^{val}_{c} \gets \mathcal{G}_{c}(D^{val}_{c})$ \Comment{Synthesizing DP data}
    \EndFor
    \State \textsc{Federated Training} \ref{alg:federatedTraining}
    
    \textit{// After collaborative training}
    \For{each client $c \in Q$ }
        \State (iii) $\hat{\theta}_{c} \gets \theta_{c} - \gamma \ast\nabla_{\theta_{c}}\mathcal{L}_{c}$
        
        $\mathcal{L}_{D^{tr}_{c}}(\theta_{c}) \gets \frac{1}{\mid D^{tr}_{c} \mid}\sum_{\left(x,y\right)\in D^{tr}_{c}}{\ell(f_{\theta_{c}}\left(x\right),\ y)}$ \Comment{Locally fine-tuning}
    \EndFor
    \end{algorithmic}
\end{algorithm}

\subsubsection{Method Overview}
We now briefly illustrate the proposed PPPFL framework, which includes 3 main phases: local data generation, federated training, and local adaptation (as depicted in Fig. \ref{fig:framework}). In our method, the DP gradient compression and aggregation approach and the optimization-based meta-learning algorithm serve as key components which help the PPPFL prevent individual information disclosures and improve the consistency among local models. 

In the first stage, to limit sensitive information leakage, each client generates its own artificial data through \textsc{DataLens}. \textsc{DataLens} \cite{wang2021datalens} is a state-of-the-art privacy-preserving data generative model, which is capable of generating high-dimensional data without leaking private information. Motivated by this novel DP-GAN model, our framework integrates it to enable each client to add noises to their own data through a training approach (Section \ref{sec:dp-training}). After this data generation step, the synthetic data ${D^*}_i$ of each $i^{th}$ client is used for the federated training, which helps to protect the original data $D_i$ as $D_i$ is kept locally by each client.

\begin{algorithm}[H]
    \caption{\textsc{Aggregations On The Server}}
    \label{alg:server}
    \begin{algorithmic}[1]
    \renewcommand{\algorithmicrequire}{\textbf{Input:}}
    \renewcommand{\algorithmicensure}{\textbf{Output:}}
    \Require initial model $\theta_0$, outer learning rate $\beta^*$
    \Ensure  aggregated model $\theta$
    \State Initialize global model $\theta \gets \theta_0$
    \For{each communication round $t=1,2,\dots$}
        \State Sample a set $U_t$ of $m$ clients and distribute $\theta_0$ for the sampled clients
        \For{each client $c \in U_t$ in parallel}
            \State $g_i \gets \textsc{Federated Training} (\theta_i)$ \ref{alg:federatedTraining}
            \State $\theta \gets \theta - \frac{\beta^*}{m}\sum_{c_t}g_i$
            \State Update the global model
        \EndFor
    \EndFor
\end{algorithmic}
\end{algorithm}

Next, during the Federated training phase (as shown in Algorithm \ref{alg:federatedTraining}), at the beginning of each communication round between clients and the server, each client downloads the weights of a global model. For every following communication round, the clients perform gradient decent updates on the synthetic training data ${D^*}^{tr}$. For the inference step, clients will get feedback by calculating the gradients on the synthetic validation dataset ${D^*}^{val}$ with respect to their own local models, and then send those feedback signals to the server for collaboration (as shown in Algorithm \ref{alg:clients}). Based on those shared signals and clients' models, the global model is updated by the server (as shown in Algorithm \ref{alg:server}). 

After the collaboration rounds stop, the clients locally conduct the adaptation phase by performing gradient descent updates on their latest model downloaded from the server with respect to the local private datasets (as shown in Algorithm \ref{alg:clients}) to get better inference performance. 

\begin{algorithm}[H]
    \caption{\textsc{Federated Training}}
    \label{alg:federatedTraining}
    \begin{algorithmic}[1]
        \renewcommand{\algorithmicrequire}{\textbf{Input:}}
        \renewcommand{\algorithmicensure}{\textbf{Output:}}
        \Require client synthetic data ${D^*}_i^{tr}, {D^*}_i^{val}$, initial model $\theta_0$, inner learning rate $\alpha^*$
        \Ensure  updated model $\theta^*$
        \State Initialize the local model $\theta \gets \theta_0$
        \For {each communication round $t=1,2,\dots$}
            \State \textit{ClientUpdate($\theta_i$, ${D^*}_i^{tr}$)}
            \State $\mathcal{L}_{D^*_{c}}^{val}(\theta_{c}) \gets \frac{1}{|D^{*val}_{c}|} \sum_{( x^{\prime}, y^{\prime})\in D^{*val}_c}{\ell(f_{\theta_i}(x^{\prime}),y^{\prime})}$ 
            \State Compute the query loss
            \State $g_i  \gets \nabla_{\theta_{i}} \mathcal{L}_{{D^*}_i^{val}} \left(\theta_i\right)$ 
            \State Uploading $g_i$ to the server
        \EndFor 
        \Function{ClientUpdate}{$\theta_i$, ${D^*}_i^{tr}$}
            \State $\mathcal{L}_{{D^*}_{c}}^{tr}\left(\theta_{c}\right)\gets\frac{1}{\left|{D^*}_{c}^{tr}\right|}\sum_{ \left( x,y \right) \in {D^*}_{c}^{tr}} \ell\left(f_{\theta_i}\left(x\right),y\right)$ 
            \State $\theta_i  \gets \theta_i - \alpha^* * \nabla_{\theta_i} \mathcal{L}_{D^{*tr}_i}$
        \EndFunction
    \end{algorithmic}
\end{algorithm}

\subsubsection{Client-side computations}
\label{sec:client-compute}

In this framework, besides the gradient sharing process with the server, the clients need to compute two extra steps, which are described in Algorithm \ref{alg:clients}. First, to ensure the instance-level DP for the framework, each client needs to train a differentially private data generator to synthesize their own dataset which is then used for collaboration. Under this setting, our framework can provide stronger privacy guarantees for individual data samples, mitigating information disclosure problems. The second additional step is that clients are expected to perform a local adaptation step on their kept private data after receiving the well-initialized parameters from the central party. This step contributes to the generation of a generalized and personalized model for each client. We assume two added steps are trivial as the clients are large organizations with high computational power. 

\subsubsection{Server-side computations}
\label{sec:server-compute}
Different from other FL settings \cite{mcmahan2017communication}, instead of naively aggregating parameters sent by clients, the objective of the server in our setting is to seek the best initialization which performs greatly after each client updates with respect to its own local private dataset. To accomplish this goal, the server aggregates shared parameters and feedback signals sent by clients $(\theta, g)$, where $g$ is the feedback signals and $\theta$ are clients' parameters. However, there always exists a trade-off between the privacy budget and utility, in which higher $\epsilon$ implies higher utility but lower privacy and vice versa. Moreover, as the $\epsilon$ value is controlled by each client, this trade-off is amplified due to the difference in clients' settings. Therefore, it would be more efficient if the server can integrate clients' privacy budget information into its optimization procedure. In other words, the objective function in PPPFL when having multiple $\epsilon$ as

\begin{equation}
    \label{eq:pppfl}
    L(\theta) = \mathbb{E}_{c \sim \mathcal{C}}\left[\frac{e^{\frac{\epsilon_{c}}{\rho}}}{\sum_{j=1}^{K} \frac{e^{\epsilon_{j}}}{\rho}} \mathcal{L}\left(\left({\theta}; {D^*}_{c}^{tr}\right); {D^*}_i^{val} \right)\right],
\end{equation}
with $\rho$ is a hyperparameter that denotes the temperature of the softmax smoothing function over the clients' budgets.



\begin{algorithm}[H]
    \caption{\textsc{DP-GANs Training} }
    \label{alg:datalens}
    \begin{algorithmic}[1]
    \renewcommand{\algorithmicrequire}{\textbf{Input:}}
    \renewcommand{\algorithmicensure}{\textbf{Output:}}
    \Require client private dataset $D_i$, $N$ discriminators $\mathcal{D}_1, \ldots, \mathcal{D}_N$
    \Ensure DP generator $\mathcal{G}$
    \State Partitioning the sensitive dataset $D_i$ into non-overlapping subsets $\mathrm{d}_{1}, \mathrm{~d}_{2}, \ldots, \mathrm{d}_{\mathrm{N}}$ an associate a teacher discriminator $\mathcal{G}_{i}$ to each set.
    \State Training the teachers: using the discrimination loss $\mathcal{L}_{\mathcal{G}}=-\log \mathcal{G}(x)-\log (1 -  \mathcal{G}\left(\mathcal{D}^{\prime}(z)\right)$ with $\mathcal{D}$ denotes the student generator $\mathrm{x}$ is sampled from the subsets $\mathrm{d}_{\mathrm{i}}, z$ is the noise sample.
    \State Gradients Compression: the teacher's dense gradients vectors $g^{(i)}$ into sparsified vectors $\tilde{g}^{(i)}$ with $\{-1,0,1\}$ values.
    \State DP Aggregating and Thresholding: the aggregated teacher's gradients vector $\tilde{g}^* \leftarrow \sum_{i=1}^N \tilde{g}^{(i)}+\mathcal{N}\left(0, \sigma^2\right)$, then be mapped to $\{-1,0,1\}$ values by comparing each gradient direction with a threshold $\beta N$. 
    \State Training the student generator: with the generator loss $\tilde{\mathcal{L}}_{\mathcal{G}}(z, \hat{x})=$ $\mathbb{E}\left[\mathcal{D}\left(z \right)-\hat{x}\right]^{2}$, where $\hat{x}=\mathcal{D}\left(\mathrm{z}\right)+\gamma \bar{g}$ is the synthetic data plus the aggregated DP teacher gradients vector and $\gamma$ denotes the learning rate.
\end{algorithmic}
\end{algorithm}

\begin{figure}[ht]
  \begin{subfigure}[b]{0.9\columnwidth}
    \includegraphics[width=\linewidth]{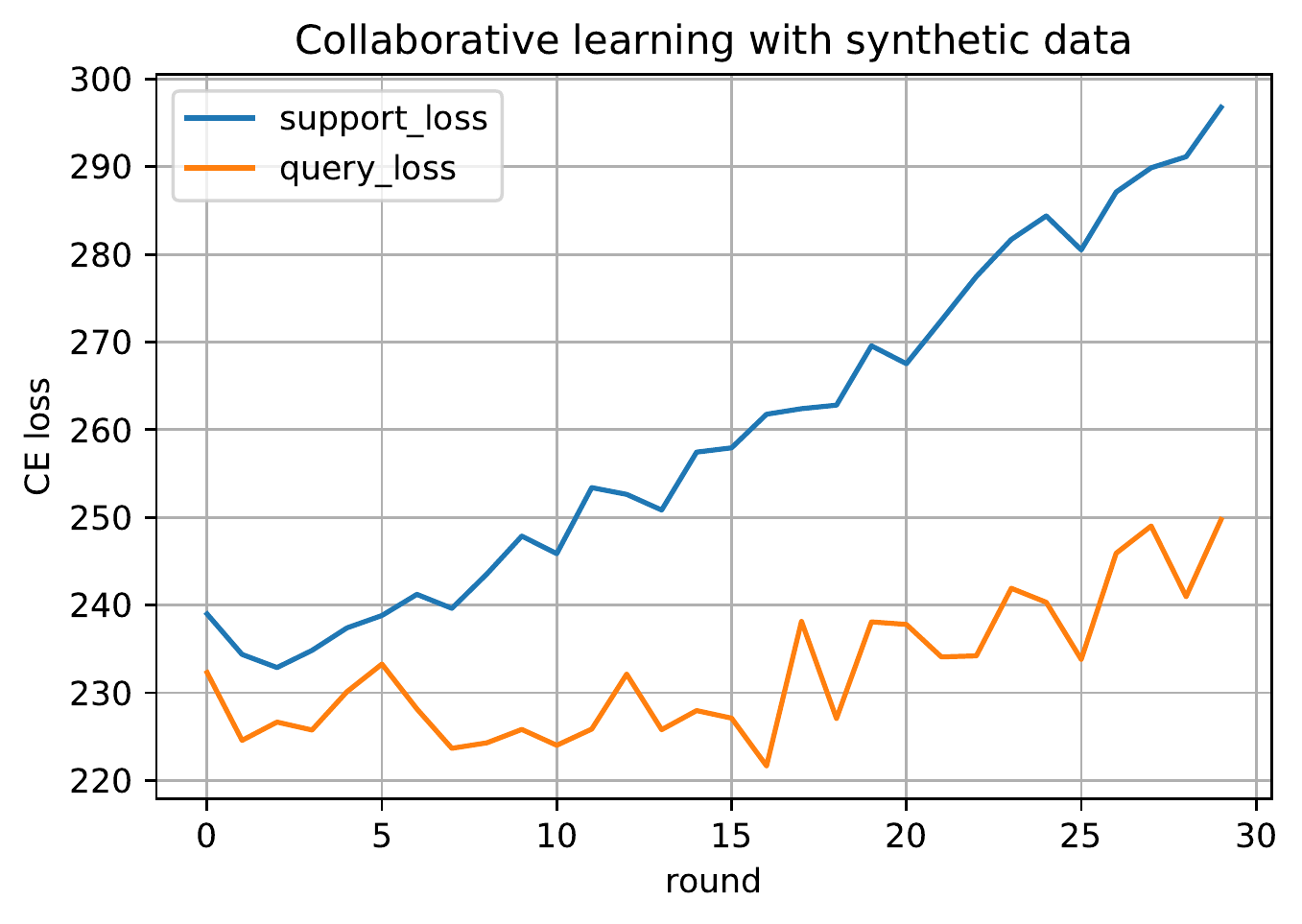}
    \caption{The diverging learning curve.}
    \label{fig:maml_before}
  \end{subfigure}
  \hspace{0.5 cm}
  \begin{subfigure}[b]{0.9\columnwidth}
    \includegraphics[width=\linewidth]{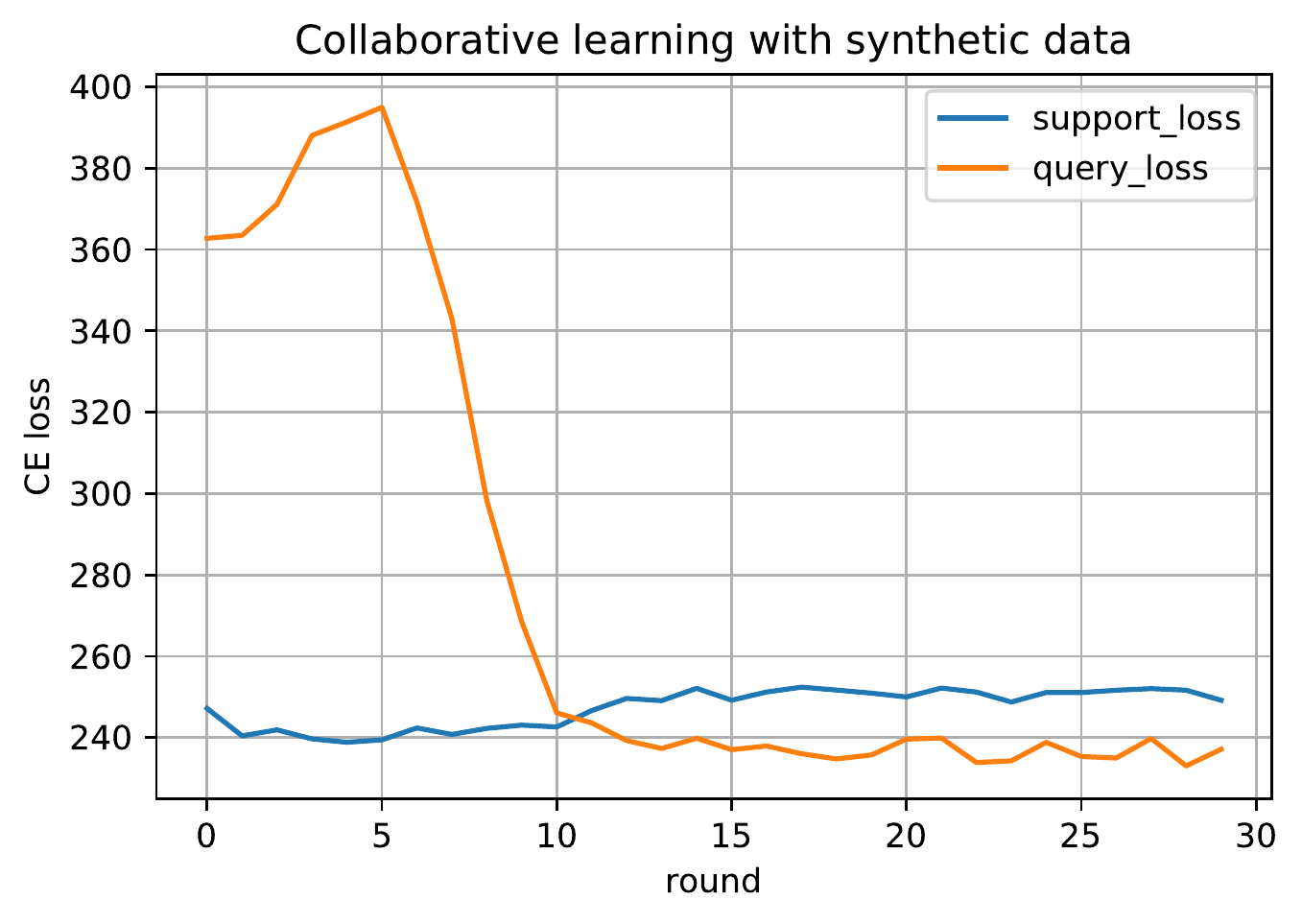}
    \caption{The stabilized learning curve.}
    \label{fig:maml_after}
  \end{subfigure}
  \caption{The improvement on convergence (bottom) compared to the original MAML (top) indicated by the total cross-entropy loss of the support set and query set over communication rounds. The top figure indicates a diverging learning curve when the gradients keep accumulating while the bottom figure illustrates fast convergence after 10 communication rounds. The experiment is conducted on the CIFAR-10 dataset.}
\end{figure}

\subsubsection{DP-GAN training}
\label{sec:dp-training}
In this paper, we consolidate the advantages of \textsc{DataLens} \cite{wang2021datalens}, which is a current state-of-the-art DP data generative model, to enable clients to generate their differential privacy dataset. Based on the PATE framework, \textsc{DataLens} consists of multiple teacher discriminators $\mathcal{D}_1, \mathcal{D}_2, \ldots, \mathcal{D}_N$ that produce gradient vectors to guide the student generator $\mathcal{G}$ in updating its synthetic samples. Meanwhile, the student generator sends updated synthetic ones to the teacher for label querying. The curated Gaussian noise $\mathcal{N}(0, \sigma)$ is added into the computations during the teacher aggregation step, which is described in Algorithm \ref{alg:datalens} and \ref{sec:privacy_analysis}. 

The detailed training procedure includes five main steps:

\begin{itemize}
    \item \textbf{Step 1: Partitioning the sensitive dataset $D_i$ into non-overlapping subsets.} This technique ensures that the discriminators do not rely on the specifics of a single sensitive training data point, hence protecting the privacy of the training data \cite{papernot2016semi}.
    \item \textbf{Step 2: Training teacher discriminators.} Each teacher discriminator $\mathcal{D}_i$ is trained to differentiate between real samples and synthetic samples generated by the student generator $\mathcal{G}$ by using the loss function $\mathcal{L}_{\mathcal{G}}=-\log \mathcal{G}(x)-\log (1 -  \mathcal{G}\left(\mathcal{D}^{\prime}(z)\right)$.
    \item \textbf{Step 3: Gradients Compression.} The gradient vector of each teacher is filtered by the ${top-k}$ values, then clipped, normalized, and lossy compressed with stochastic sign gradient quantization to reduce communication cost and to improve privacy and convergence guarantees.
    \item \textbf{Step 4: DP gradient aggregation and thresholding.} The Gaussian noise is calibrated by estimating $\ell_2$ sensitivity of the aggregated gradient vector (as shown in \ref{sec:privacy_analysis}). The 
    \item \textbf{Step 5: Training the student generator.} When gradient descent steps are performed on the loss function of the student generator, the aggregated DP gradient vectors from the teacher discriminators are backpropagated to the student generator.
\end{itemize}

\subsubsection{Stabilizing MAML algorithm}
As previously noted by \citeauthor{antoniou2018train} \cite{antoniou2018train}, the optimization process in Model-Agnostic Meta-Learning (MAML) can be unstable due to the multiple times that differentiation is performed through the model. This instability is illustrated in Fig. \ref{fig:maml_before}, which shows that the original MAML model diverges after just a few communication rounds. To address this issue, we propose a modified version of MAML that employs Model Exponential Moving Average in the inner loop, followed by Cosine Annealing Learning rate scheduling in the outer loop. This approach is inspired by previous work on improving generalization in DL over Stochastic Gradient Descent (SGD) in general \cite{izmailov2018averaging, tarvainen2017mean, huang2017snapshot} and improving stability in MAML in specific \cite{antoniou2018train}. 
Formally, the model parameters $\theta_t$ of the clients at communication round $t$ are set to be an Exponential Moving Average (EMA) of the previous round's model parameters: $\theta_t := \zeta \theta_t + (1-\zeta) \theta_{t-1}$, where $\zeta$ is the momentum hyperparameter. 
The learning rate $\beta^{*}$ for the outer optimization loop is calculated as $\beta^*(t)=f(\bmod (t-1,\lceil T / M\rceil))$, where $t$ is the current round of the total $T$ communication rounds, $M$ is the number of cosine annealing cycles, and $f$ is a monotonically decreasing function. By combining these techniques, the global model is able to traverse multiple local optima while maintaining stability during training.

\subsubsection{Personalization} 

In our approach, we also take the customization and personalization of clients into consideration since they may have different business-related requirements for data sharing. Instead of giving the server full control on the level of user privacy, we allow clients to adjust their own privacy level depending on their actual demands. This enables clients to independently, and effectively select their preferred privacy budgets $\epsilon$ and incorporate them directly into the DP-GAN training procedure while still being able to control the outcomes of the training process without having to wait for the collaborative model to converge. Additionally, the information about the client's privacy budgets is used in our proposed framework to improve the performance of the client's models.

\subsection{Privacy Analysis
}
\label{sec:privacy_analysis}
In our chosen DP-generative algorithm \cite{wang2021datalens}, briefly described in Algorithm \ref{alg:datalens}, the gradients of the student generator are obtained by aggregating the sparsified gradients of the teachers and voting. Specifically, the gradient set $G=\{g^{(1)},\ldots,g^{(\mathrm{N})} \}$ of $N$ is quantized using top-$k$-sparsification \cite{shi2019understanding} to convert the real-valued gradient vector into a sparse $\{-1, 1\}$ vector $\tilde{g}^{(i)}$ with $k$ non-zero entries.

The Gaussian mechanism can be applied to the sum aggregation function defined as $f_{\text{sum}}(\tilde{\mathcal{G}})=\sum_{i=1}^N \tilde{\mathbf{g}}^{(i)}$ for a set of compressed teacher gradient vectors $\tilde{G}=\left(\tilde{\mathbf{g}}^{(1)}, \ldots, \tilde{\mathbf{g}}^{(N)}\right)$, where $\tilde{\mathbf{g}}^{(i)}$ is the compressed gradient of the $i$-th teacher. The resulting output is $\tilde{\mathbf{G}}\sigma f{\text{sum}}(\tilde{\mathcal{G}})=f_{\text{sum}}(\tilde{\mathcal{G}})+\mathcal{N}\left(0, \sigma^2\right)=\sum_{\tilde{\mathbf{g}} \in \tilde{\mathcal{G}}} \tilde{\mathbf{g}}+\mathcal{N}\left(0, \sigma^2\right)$. The Gaussian mechanism guarantees $(\lambda, \alpha)$-RDP \ref{def:rdp} for any real-valued function $f$. 

\begin{lemma} For any neighboring top- $k$ gradient vector sets $\tilde{\mathcal{G}}, \tilde{\mathcal{G}}^{\prime}$ differing by the gradient vector of one teacher, the $\ell_2$ sensitivity for $f_{\text {sum }}$ is $2 \sqrt{k}$.
\end{lemma}

According to Theorem 1, the Gaussian mechanism $\mathrm{G}\sigma f$ to a function $f$ with $\ell_2$ sensitivity results in a $(\lambda, s^2 \lambda /(2 \sigma^2))$-RDP guarantee. 
In order to determine the RDP guarantee for the Gaussian mechanism applied to the sum aggregation function $\tilde{\mathrm{G}}\sigma f_{\mathrm{sum}}(\tilde{\mathcal{G}})$, it is necessary to calculate the $\ell_2$ sensitivity of this function. According to Lemma 1, the $\ell_2$ sensitivity of $f_{\text{sum}}$ for any neighboring top-$k$ gradient vector sets $\tilde{\mathcal{G}}, \tilde{\mathcal{G}}^{\prime}$ that differ by a single gradient vector is equal to $2 \sqrt{k}$. This value can be used to determine the RDP guarantee for the Gaussian mechanism applied to the sum aggregation function.

\begin{table*}[ht]
\centering
\caption{The average performance over the clients, trained on synthetic, fine-tuned and test on secret data.}
    \begin{tabular}{l|cc|cc|cc|cc}
    \multicolumn{1}{c|}{} & \multicolumn{2}{c|}{\textbf{MNIST}} & \multicolumn{2}{c|}{\textbf{FMNIST}} & \multicolumn{2}{c|}{\textbf{CIFAR-10}} & \multicolumn{2}{c}{\textbf{CIFAR-100}} \\ \hline
    Method & \multicolumn{1}{l}{BMT-F1} & \multicolumn{1}{l|}{BMTA} & \multicolumn{1}{l}{BMT-F1} & \multicolumn{1}{l|}{BMTA} & \multicolumn{1}{l}{BMT-F1} & \multicolumn{1}{l|}{BMTA} & \multicolumn{1}{l}{BMT-F1} & \multicolumn{1}{l}{BMTA} \\ \hline
    FedAvg \cite{mcmahan2017communication} & 0.9223 & 0.9051 & 0.7267 & 0.7938 & 0.4898 & 0.6562 & 0.2142 & 0.3028 \\
    FedNova \cite{wang2020tackling} & 0.8773 & 0.8996 & 0.6303 & 0.7607 & 0.3847 & 0.6627 & 0.2855 & 0.3374 \\
    FedProx \cite{li2020federated} & 0.9006 & 0.9040 & 0.6994 & 0.7721 & 0.4856 & 0.6718 & 0.3014 & 0.3404 \\
    SCAFFOLD \cite{karimireddy2020scaffold} & 0.8906 & 0.9024 & 0.6698 & 0.8061 & 0.3732 & 0.6614 & 0.2096 & 0.3333 \\
    FedMeta \cite{chen2018federated} & 0.9399 & 0.9616 & 0.8367 & 0.8800 & 0.6897 & 0.7466 & 0.2737 & 0.4133 \\
    \textbf{PPPFL (ours)} & \textbf{0.9420} & \textbf{0.9640} & \textbf{0.8622} & \textbf{0.9199} & \textbf{0.7163} & \textbf{0.8000} & \textbf{0.3526} & \textbf{0.4800} \\ \hline
    \end{tabular}%
\label{tab:main_results}
\end{table*}

\section{Evaluation}
\label{sec:evaluation}
In this section, we illustrate the experimental evaluation of our proposed framework for two problems: prevent sensitive information leakage and address clients' data heterogeneity issues. To demonstrate its effectiveness, we conduct experiments on different benchmark datasets and compare them with six baseline models.

\begin{figure}[!ht]
    \centering
    \includegraphics[width=0.95\linewidth]{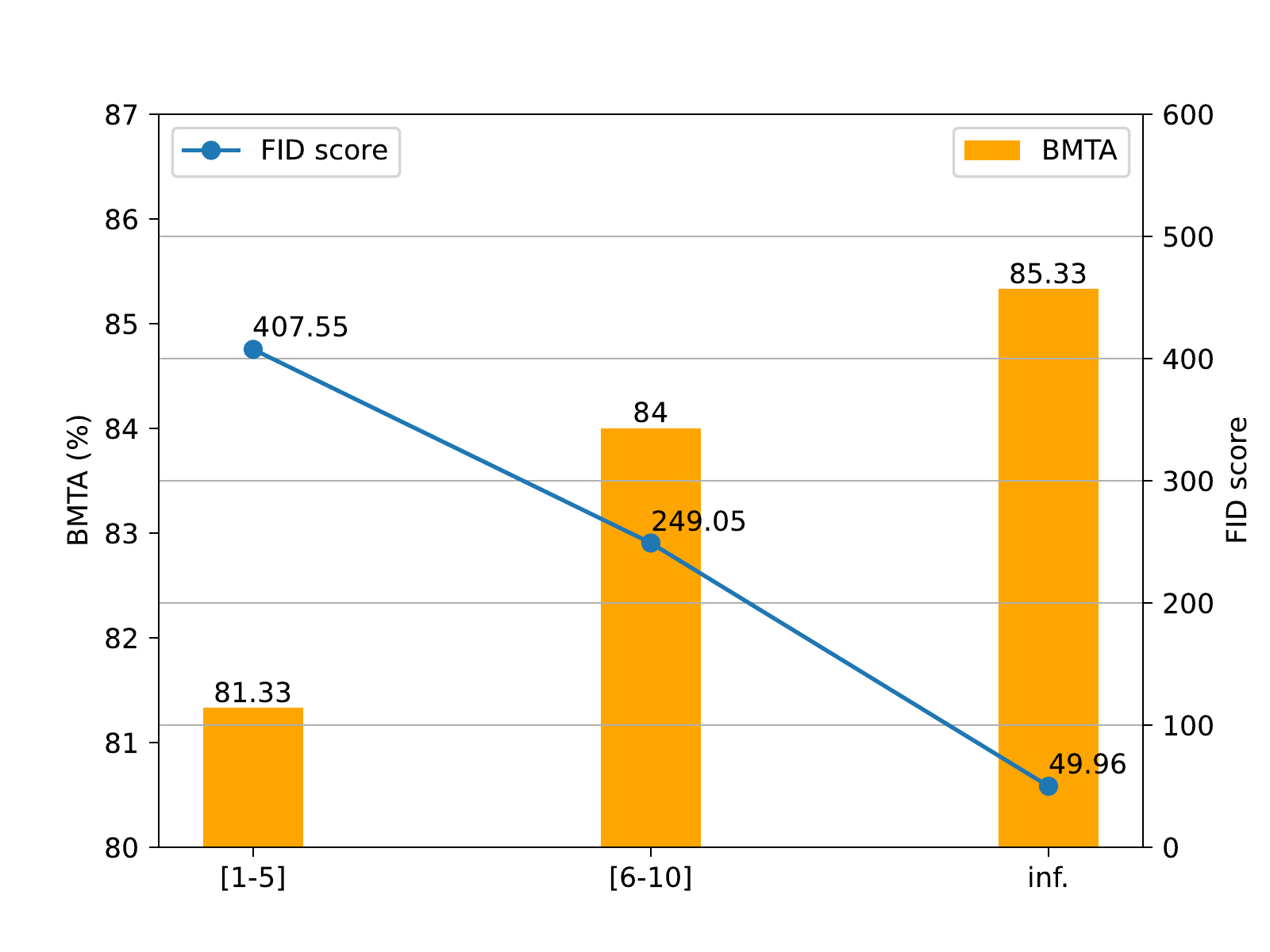}
    \caption{Test accuracy and FID score with respect to different users' privacy budgets. The left vertical axis represents the BMTA in percentage, the right vertical axis indicates the synthetic data quality in terms of FID score while the categories in the horizontal axis are different ranges of clients' privacy budget $\epsilon$, specifically from $1$ to $5$, $6$ to $10$, and lastly $\texttt{inf.}$ represents synthetic data without any DP guarantees.}
    \label{fig:ablation_synthetic_chart}
\end{figure}

\subsection{Experiment Setup}
\noindent

\begin{table}[ht]
    \centering
    \caption{The BMTA of adapted models when using non-private GANs}
    \resizebox{0.95 \columnwidth}{!}{%
    \begin{tabular}{l|c|c|c}
    Method         & \textbf{MNIST} & \textbf{FMNIST} & \textbf{CIFAR-10} \\ \hline
    FedAvg \cite{mcmahan2017communication}        & 0.9400                                           & 0.7893        & 0.5521               \\
    FedNova \cite{wang2020tackling}       & 0.8386                                          & 0.7782         & 0.5724               \\
    FedProx \cite{li2020federated}       & 0.9373                                          & 0.7981           & 0.5737                \\
    SCAFFOLD \cite{karimireddy2020scaffold}      & 0.9279                                          & 0.8061           & 0.5608             \\
    SDA-FL \cite{huang2021personalized}     & \textbf{0.9826}                                & \textbf{0.8687}  & 0.6789                       \\
    \textbf{PPPFL (ours)}          & 0.9466                                 & 0.8666    & \textbf{0.7133}   \\
    \hline
    \end{tabular}
}
    \label{tab:ablation_non_private_gans}
\end{table}
\noindent
\textbf{Datasets}. We use four public benchmark datasets, including MNIST \cite{deng2012mnist}, Fashion-MNIST \cite{xiao2017fashion}, CIFAR-10 \cite{krizhevsky2010convolutional}, and CIFAR-100, to evaluate our proposed method. MNIST and Fashion-MNIST have $60000$ training examples and $10000$ testing examples, and all of them are grayscale images of $28 \times 28$ dimensions. We also utilize two color-image datasets, CIFAR-10 and CIFAR-100, to demonstrate the performance of the proposed method. Both datasets have $50000$ in training images and $10000$ in testing images.

\noindent
\textbf{Baselines}. We compare our proposed method with the most popular FL algorithm FedAvg \cite{mcmahan2017communication}, and state-of-the-art FL algorithms under non-IID data, including FedNova \cite{wang2020tackling}, FedProx \cite{li2020federated}, SCAFFOLD \cite{karimireddy2020scaffold}, FedMeta \cite{chen2018federated}, and SDA-FL \cite{huang2021personalized}. 

\noindent
\textbf{Federated Learning Settings.} We simulate the FL scenario with $5$ clients and $30$ communication rounds. In each communication round, all of the clients participate in the protocol. The inner learning rate and outer learning rate for the meta-training procedure are respectively set to $1e-4$ and $3e-4$, the used optimizer is Adam \cite{kingma2014adam} and batch size is $64$.

\begin{figure}[!htbp]
\centering
  \label{fig:incentive_analysis}
  \begin{subfigure}[b]{0.85\columnwidth}
    \includegraphics[width=\linewidth]{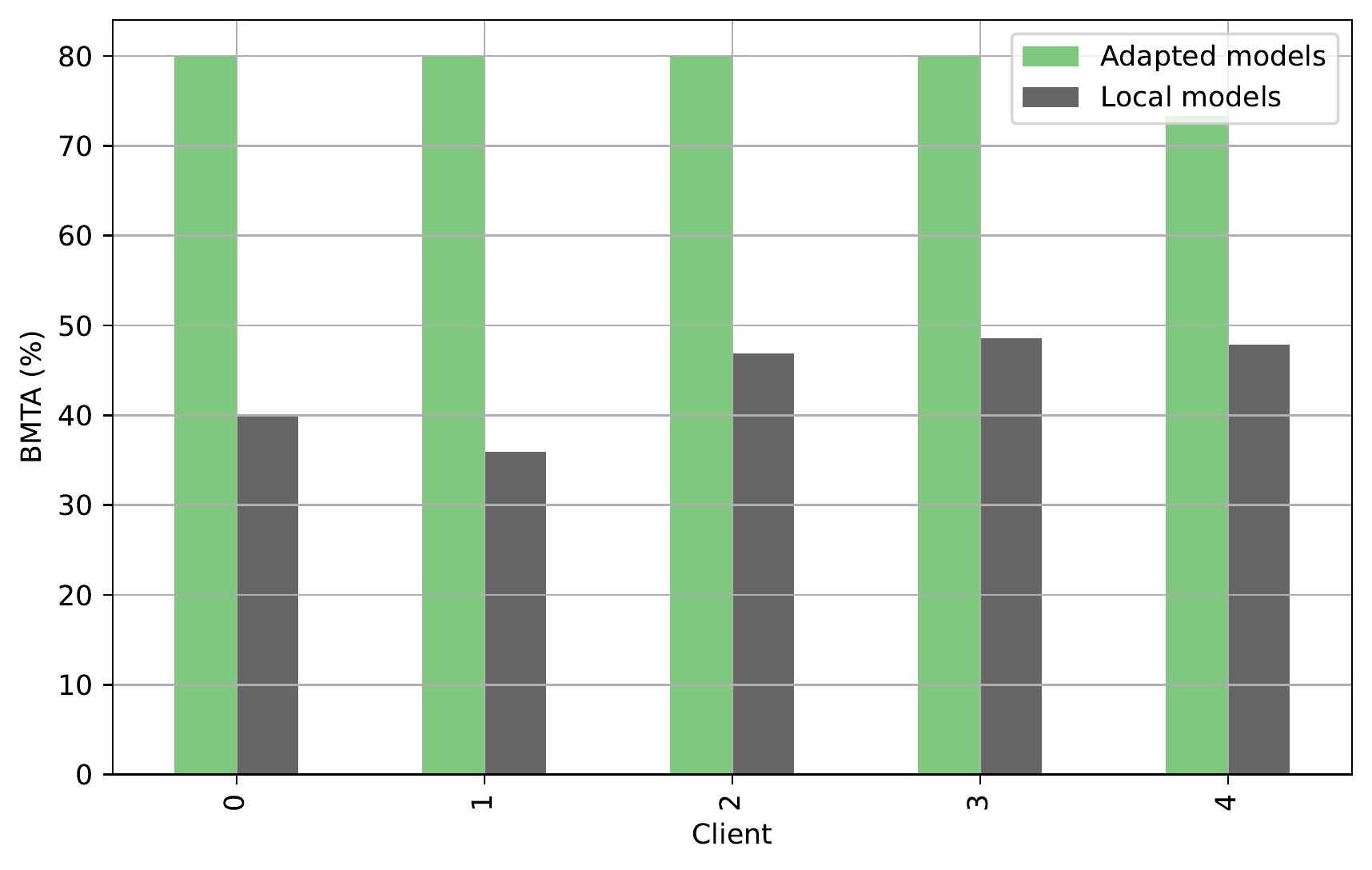}
    \caption{Accuracy improvements of adapted federated models over local, trained-from-scratch models.}
    \label{fig:incentive_changes}
  \end{subfigure}
  \hspace{0.5 cm}
  \begin{subfigure}[b]{0.85\columnwidth}
    \includegraphics[width=\linewidth]{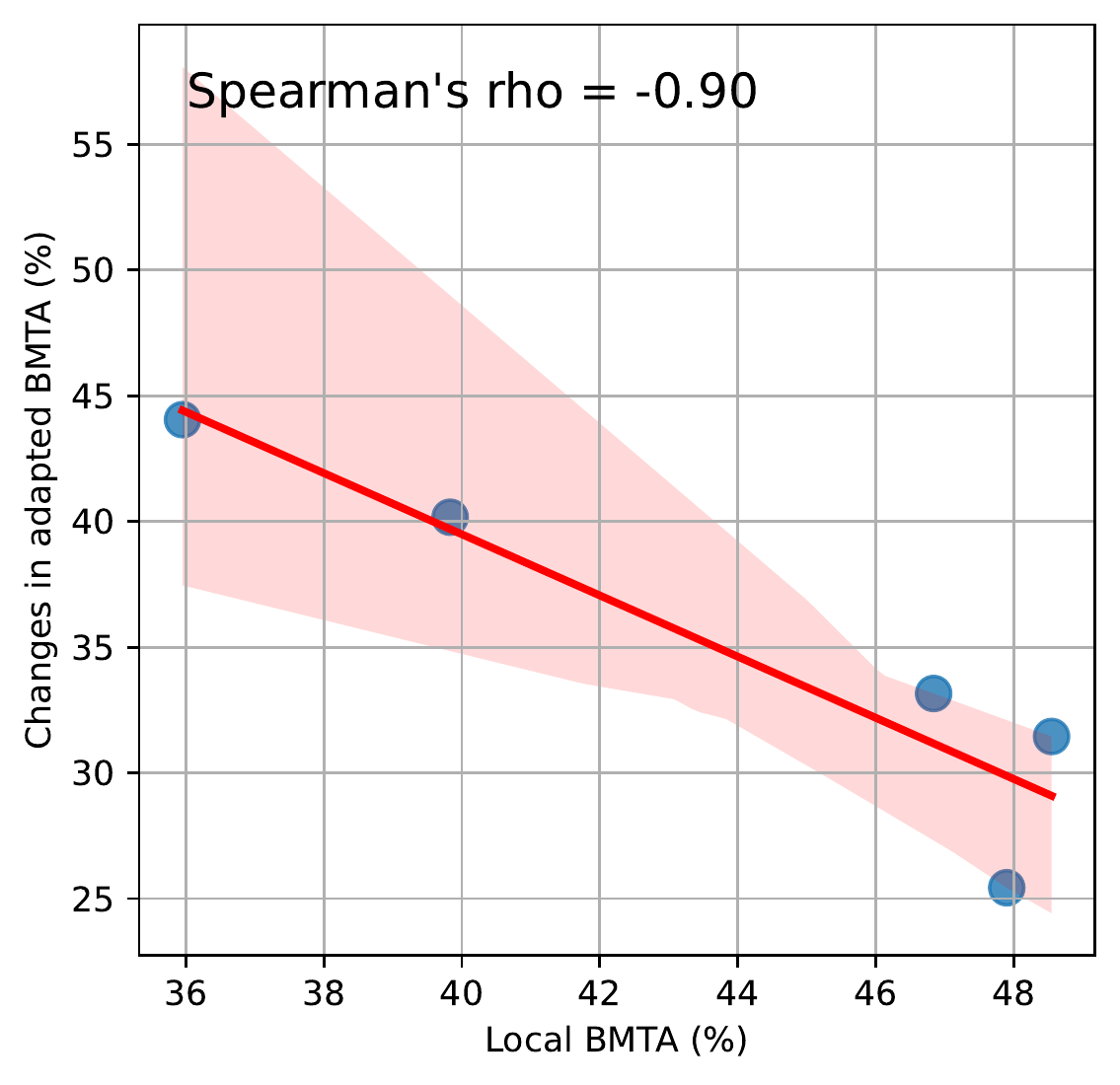}
    \caption{The correlation between local models' performance with the performance gain from protocol participation.}
    \label{fig:incentive_changes_correlation}
  \end{subfigure}
  \caption{The relationship between the enhancement of adapted federated models and locally trained models. The Spearman's correlation coefficient is  $ -0.90$ indicating a strong correlation between local models' performance with the performance gained by taking part in the protocol.}
\end{figure}

\begin{table*}[ht]
\begin{threeparttable}
\caption{Additional experiments on different training and testing setups\tnote{1}}
\label{tab:ablation_training_test_setup}
\begin{tabular}{c|l|cc|cc|cc|cc}
\multirow{2}{*}{\textbf{Method}} & \multicolumn{1}{c|}{\multirow{2}{*}{\textbf{Train / test setup}}} & \multicolumn{2}{c|}{\textbf{MNIST}} & \multicolumn{2}{c|}{\textbf{FMNIST}} & \multicolumn{2}{c|}{\textbf{CIFAR-10}} & \multicolumn{2}{c}{\textbf{CIFAR-100}} \\ \cline{3-10} 
 & \multicolumn{1}{c|}{} & \multicolumn{1}{l}{BMT-F1} & \multicolumn{1}{l|}{BMTA} & \multicolumn{1}{l}{BMT-F1} & \multicolumn{1}{l|}{BMTA} & \multicolumn{1}{l}{BMT-F1} & \multicolumn{1}{l|}{BMTA} & \multicolumn{1}{l}{BMT-F1} & \multicolumn{1}{l}{BMTA} \\ \hline
\multirow{2}{*}{\textbf{Local}} & secret / secret & 0.8292 & 0.8412 & 0.6896 & 0.7136 & 0.5959 & 0.6130 & 0.2721 & 0.2998 \\
 & synthetic / secret & 0.3021 & 0.3371 & 0.0716 & 0.0838 & 0.1450 & 0.1578 & 0.0067 & 0.0100 \\ \hline
\multirow{3}{*}{\textbf{Collaborative}} & synthetic / secret & 0.2689 & 0.3157 & 0.0476 & 0.1231 & 0.0500 & 0.1417 & 0.0089 & 0.0100 \\
 & synthetic - fine-tune / secret\tnote{2} & 0.9223 & 0.9051 & 0.7267 & 0.7938 & 0.5510 & 0.7124 & 0.2142 & 0.3028 \\
 & secret / secret & 0.9360 & 0.9645 & 0.8730 & 0.8972 & 0.7427 & 0.8083 & 0.4520 & 0.5416 \\ \hline
\end{tabular}
\begin{tablenotes}
    \item[1] The "Train/test setup" column in the table specifies the data used for training and testing the models for each method. The format of the entries in this column is "train set/test set", where the "train set" is the data used to train the model and the "test set" is the data used to evaluate the model's performance.
    \item[2] refers to a training and testing setup in which the model is first trained on synthetic data, then fine-tuned on real, secret data, and finally evaluated on a real, secret test set.
\end{tablenotes}
\end{threeparttable}
\end{table*}

\subsection{Performance on DP-synthetic data}

\noindent
\textbf{DP data generation model.}
We follow the hyperparameter choices by \cite{wang2021datalens}. Specifically, for the MNIST Handwritten Dataset, the DP GANs are trained using privacy budget $\epsilon \in { 1, \dots, 10} $. The sensitivity $\delta$ is set to $1e-5$, $top-k$ value equals to $300$. The noise parameter $\sigma = 5000$ and voting threshold $\beta = 0.7$.
\\
\noindent
\textbf{Non-IID simulation.}
We simulate the common non-IID scenario where the label distribution is skewed. The label distribution $q \sim Dirichlet(\tau)$ with $\tau = 0.5$.
\\
\noindent
\textbf{Metrics.}
\noindent
We follow the work by \citeauthor{huang2021personalized} \cite{huang2021personalized} on employing Best Mean Test Accuracy (BMTA) as the metric by averaging the testing accuracy on all clients and selecting the best value across the training communication rounds. In cases of the presence of non-IID data, we use the Best Mean Test F1-score (BMT-F1) as an additional metric for better evaluation. Formally, at communication round $t$, given the accuracy score $\nu_i^t$ and the F1-score $\zeta_i^t$ of the $o^{th}$ client ($i \in [1, K]$), the BMTA and BMT-F1 metrics are calculated as $\text{BMTA} = \max_{{t}} \frac{1}{K} \sum_{i=1}^{K} \nu_i^t$ and $\text{BMT-F1} = \max_{{t}} \frac{1}{K} \sum_{i=1}^{K} \zeta_i^t$.

Table \ref{tab:main_results} shows that our proposed framework PPPFL outperforms the baselines by a significant margin. Specifically, we outperform the baselines by a significant margin in multiple datasets, we also push the frontier of data privacy and data utility by achieving good performance while maintaining DP guarantees. In terms of accuracy, our proposed method outperforms previous works in all four benchmarks, MNIST, FMNIST, CIFAR-10, and CIFAR-100, with the corresponding figures: 0.9640, 0.9199, 0.80, and 0.480. Regarding BMT-F1, in the CIFAR-10 dataset, our framework is also superior to the FedMeta algorithms by more than 2.5\%, while this figure achieves by nearly 8.0\% regarding the CIFAR-100 dataset.

\subsection{Performance on non-private synthetic data.}
Table \ref{tab:ablation_non_private_gans} shows that when using only a single global model, in comparison with the ensemble of multiple global models in SDA-FL \cite{li2022federated}, our proposed framework can achieve comparable performances on MNIST and FMNIST datasets and obtain higher accuracy on CIFAR-100 dataset, at 0.7133.

\subsection{Performance on other data flows.}
Table \ref{tab:ablation_training_test_setup} is divided into two main categories, including the first one is the first two rows that show the results of models trained locally, and the second one is the last three rows that illustrate collaboration results between these models. The purpose of this experiment is to approximate the lower bounds and upper bounds of the models' performance when using different training and testing data setups. Regarding the first group, the independent local training scenario without FL is conducted to form a strong baseline. For the second group, FL collaboration procedures with and without using private data as the clients' training sets are performed to demonstrate the necessity of local adaptation. 


\subsection{Trade-off between Synthetic Data Privacy Levels and Model Performance} 
We evaluate our proposed framework under different users' data privacy budget $\epsilon$, from the low range $\epsilon 
\in [1 \ldots 5] $ to a higher range between $[6, \ldots, 10]$, and lastly a scenario where the synthetic data is generated by a generative model without having any added DP-noise. The quality of those aforementioned synthetic data sets is evaluated using Inception Score (IS) and Frechet Inception Distance (FID) \cite{heusel2017gans}. The results shown in Table \ref{tab:ablation_synthetic_quality} and Fig. \ref{fig:ablation_synthetic_chart} indicate that the higher DP-noise added in the synthetic data might result in higher generalization error.

\begin{table}[H]
\caption{Test accuracy and Macro F1-score with respected to synthetic data quality.}

\begin{tabular}{l|c|c|c|c}
                             & \textbf{BMTA} $\uparrow$ & \textbf{BMT-F1} $\uparrow$ & \textbf{IS} $\uparrow$ & \textbf{FID} $\downarrow$ \\ \hline
$\epsilon \in [1 \ldots 5]$  & 0.8133                 & 0.7454                & 1.2242          & 407.5517           \\
$\epsilon \in [6 \ldots 10]$ & 0.8400                 & 0.7940                & 1.8100          & 249.0475           \\
$\epsilon = \infty$          & 0.8533                 & 0.8199                & 2.9442          & 49.9617            \\ \hline
\end{tabular}

\label{tab:ablation_synthetic_quality}
\end{table}

\section{Conclusions and Future Works}
\label{sec:discussion}

In this work, we show that our proposed Personalized Privacy-Preserving Federated Learning (PPPFL) framework could not only minimize the risk of sensitive information being leaked but also overcome the challenges of the nonindependent and identically distributed (non-IID) data among clients. We demonstrate the efficiency of the PPPFL framework as it outperforms many existing baselines on both DP-synthetic and non-private synthetic data when evaluating different datasets, including MNIST, Fashion-MNIST, CIFAR-10, and CIFAR-100. We also indicate the existence of a trade-off between synthetic data privacy levels and the model's performance. Despite our framework's performance improvements, conducting experiments with other GANs architectures is necessary.

\textbf{User incentive to participate in the protocol.} In this part of our work, we focus on the incentives for users to participate in the PPPFL framework. In cross-silo FL, organizations or companies usually have intentionally long-term strategic focuses and development goals. This makes the long-term cooperation relationship more possible, as long as we can design a proper incentive mechanism. To this end, we demonstrate a strong correlation between the local performance of a user's model and the benefits that each client derives from participating in the protocol. This correlation is shown in Fig. \ref{fig:incentive_changes_correlation}. Particularly, if local models have lower performance, it is evident that they would gain more from participating in the PPPFL protocol.

This phenomenon raises the possibility of fairness issues within the PPPFL framework, which we plan to address in future work.


%

\ifCLASSOPTIONcompsoc
  \section*{Acknowledgments}
\else
  \section*{Acknowledgment}
\fi

This work was supported by VinUni-Illinois Smart Health Center (VISHC), VinUniversity.

\ifCLASSOPTIONcaptionsoff
  \newpage
\fi




%
\bibliography{references}

%

\begin{IEEEbiography}[{\includegraphics[width=1in,height=1.25in,clip,keepaspectratio]{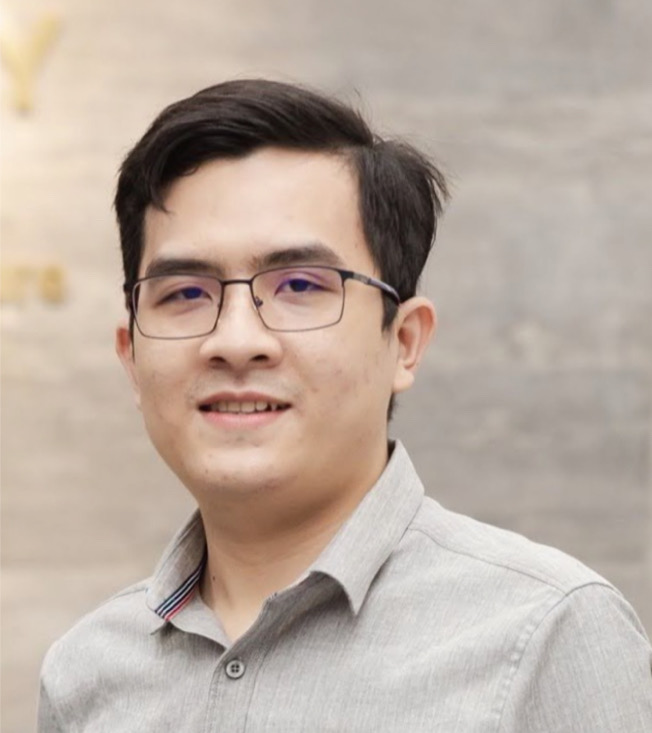}}]{Van-Tuan Tran} got the Bachelor Degree of Engineering in Telecommunications Engineering from Vietnam National University, Ho Chi Minh City. He was a Research Intern at Smart Health Center, VinBigData JSC, and a Research Assistant at VinUni-Illinois Smart Health Center, VinUniversity, Hanoi, Vietnam. His research interests include Deep Learning Applications for Computer Vision, Secure AI and Federated Learning. 
\end{IEEEbiography}



\begin{IEEEbiography}[{\includegraphics[width=1in,height=1.25in,clip,keepaspectratio]{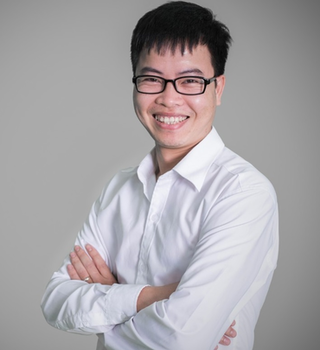}}]{Hieu H. Pham}
(Member, IEEE) received the
Engineering degree in industrial informatics from
the Hanoi University of Science and Technology
(HUST), Vietnam, in 2016, and the Ph.D. degree
in computer science from the Toulouse Computer
Science Research Institute (IRIT), University of
Toulouse, France, in 2019. He is currently a
Assistant Professor at the College of Engineering
and Computer Science (CECS), VinUniversity,
and serves as an Associate Director at the VinUni-Illinois Smart Health Center (VISHC). His research interests include computer vision,
machine learning, medical image analysis, and their applications in smart
healthcare. He is the author and coauthor of 30 scientific articles appeared
in about 20 conferences and journals, such as Scientific Data (Nature), Computer Vision and
Image Understanding, Neurocomputing, International Conference on Medical Image Computing and Computer-Assisted Intervention (MICCAI), Medical Imaging with Deep Learning (MIDL), IEEE International Conference on
Image Processing (ICIP), and IEEE International Conference on Computer
Vision (ICCV). He is also currently serving as Reviewer for MICCAI,
ICCV, CVPR, IET Computer Vision Journal (IET-CVI), IEEE Journal of Biomedical and Health Informatics (JBHI), and Scientific Reports (Nature). Before joining VinUniversity, he worked at the Vingroup Big Data Institute (VinBigData), as a Research Scientist and the Head of the Fundamental
Research Team. With this position, he led several research projects on
medical AI, including collecting various types of medical data, managing
and annotating data, and developing new AI solutions for medical analysis.
\end{IEEEbiography}


\begin{IEEEbiography}[{\includegraphics[width=1in,height=1.25in,clip,keepaspectratio]{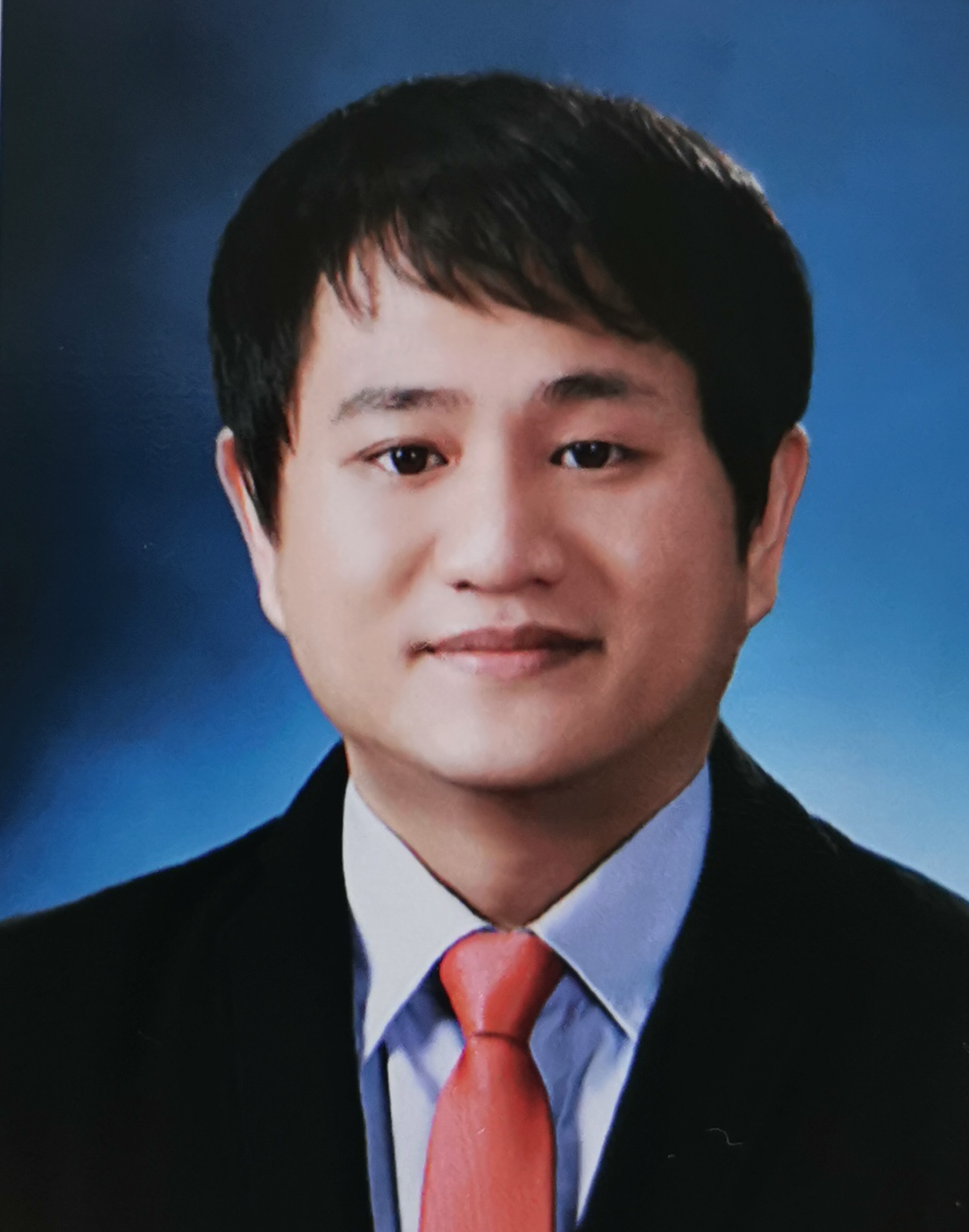}}]{Kok-Seng Wong} (Member, IEEE) received his first degree in Computer Science (Software Engineering) from the University of Malaya, Malaysia in 2002,
and an M.Sc. (Information Technology) degree from the Malaysia University of Science and Technology (in collaboration with MIT) in 2004. He obtained his Ph.D. from Soongsil University, South Korea,
in 2012. He is currently an Associate Professor in the College of Engineering and Computer Science, VinUniversity. Before joining VinUniversity, he taught computer science subjects (undergraduate
and postgraduate) in Kazakhstan, Malaysian, and South Korean universities
for the past 17 years. Dr. Wong aims to bring principles and techniques from cryptography to the design and implementation of secure and privacy-protected systems. He has published 60 articles in journals and conferences
in his research field. To this end, he conducts research that spans the areas of
security, data privacy, and AI security while maintaining a strong relevance
to the privacy-preserving framework.
\end{IEEEbiography}

\end{document}